\documentclass{article}

\usepackage{arxiv}
\usepackage{mathtools}
\usepackage[utf8]{inputenc} 
\usepackage[mathcal]{eucal}
\usepackage[T1]{fontenc}    
\usepackage{hyperref}       
\usepackage{url}            
\usepackage{booktabs}       
\usepackage{amsfonts}       
\usepackage{amsmath}
\usepackage{nicefrac}       
\usepackage{microtype}      
\usepackage{lipsum}
\usepackage{fancyhdr}       
\usepackage{graphicx}       
\usepackage[ruled,vlined]{algorithm2e}  
\graphicspath{{media/}}     

\title{ABAW : Facial Expression Recognition in the wild}

\author{
  Dr. Darshan Gera \\
  SSSIHL, Brindavan Campus \\
  Bengaluru, Karnataka, India\\
  \texttt{darshangera@sssihl.edu.in} \\
  \And
  Badveeti Naveen Siva Kumar\\
  SSSIHL, Prasanthi Nilayam Campus \\
  Sri Sathya Sai District, Andhra Pradesh, India \\
  \texttt{bnaveensivakumar@gmail.com} \\
  \AND
 Bobbili Veerendra Raj Kumar\\
  SSSIHL, Prasanthi Nilayam Campus \\
  Sri Sathya Sai District, Andhra Pradesh, India \\
  \texttt{veerendra.rajkumar@gmail.com} \\
  \And
  Dr. S Balasubramanian \\
  SSSIHL, Prasanthi Nilayam Campus \\
  Sri Sathya Sai District, Andhra Pradesh, India\\
  \texttt{sbalasubramanian@sssihl.edu.in} \\
}

\hypersetup{colorlinks=true, urlcolor=cyan,
pdftitle={Multi Task Learning for Affect Recognition at ABAW Competition Using Semi Supervised Learning},
pdfauthor={Darshan Gera, S Balasubramanian, Badveeti Naveen Siva Kumar, Bobbili Veerendra Raj Kumar},
pdfkeywords={Multi Task Learning, Semi Supervised Learning, Facial Expression Recognition, AffWild2, s-aff-wild2},
}

\begin{document}
\maketitle

\begin{abstract}
The fifth Affective Behavior Analysis in-the-wild (ABAW) competition has multiple challenges such as Valence-Arousal Estimation Challenge, Expression Classification Challenge, Action Unit Detection Challenge, Emotional Reaction Intensity Estimation Challenge. In this paper we have dealt only expression classification challenge using multiple approaches such as fully supervised, semi-supervised and noisy label approach. Our approach using noise aware model has performed better than baseline model by 10.46\% and semi supervised model has performed better than baseline model by 9.38\% and the fully supervised model has performed better than the baseline by 9.34\%.
\end{abstract}


\section{Introduction}
Facial expression recognition (FER) is also a rapidly growing field of research that has become increasingly important in recent years. The ability to accurately detect and interpret human emotions based on facial expressions has a wide range of potential applications, from improving human-computer interaction to enhancing mental health diagnosis and treatment. To get models that are independent of demographic features such as age, gender, region, we need to train the models on real-in-wild datasets such as RAF-DB, Affectnet, Aff-wild2 {\cite{kollias2017recognition,kollias2018aff,kollias2018multi,kollias2018photorealistic,kollias2019deep,kollias2019expression,kollias2020va,kollias2019face,kollias2020deep,kollias2020analysing,kollias2023abaw,kollias2021affect,kollias2022abawECCV,kollias2021analysing,kollias2021distribution,kollias2022abawCVPR,zafeiriou2017affwild_dataset}}. These in-wild datasets poses multiple challenges such as variation in illumination, variation in poses, blur, etc. 
In this paper we have dealt with this problem in multiple approaches such as fully supervised,semi-supervised and noisy label approach.

\section{Method}
In this section, we present our solution to the Expression (Expr) Classification Challenge at the 5th Affective Behavior Analysis in-the-wild (ABAW) Competition. 
\subsection{Baseline} \label{sec:baseline}
The white paper\cite{kollias2023abaw} released by organisers of ABAW competition provided as baseline, a VGG16 network with fixed convolutional weights from a pre-trained checkpoint on VGGFACE dataset for feature extraction and the last three fully connected layers are trainable along with output layer equipped with softmax activation function that gives the predictions on 8 expression classes. 

\subsection{Fully Supervised approach with finetuning} \label{sec:fully_supervised}
The approach taken in the baseline model of \cite{kollias2023abaw} is to use a pre-trained checkpoint as a fixed feature extractor. We improve this by using resnet-18 \cite{he2016deepresnet} as basenet with pre-trained weights from resnet-18 model trained on MS-Celeb \cite{msceleb} face recognition data-set as our initial point, which acts as feature extractor, this is fine-tuned on the current task of facial expression recognition. This basenet is followed by a dropout layer and fully connected layer equipped with softmax activation learns to predict on the 8 expression classes considered as part of this challenge.

To train this model, images were augmented using augmentations such as horizontal flip, random crop and were resized to 224*224*3 before feeding them to the model to obtain predictions. Cross entropy $^{\ref{eq:lossCE}}$ was used as loss function for the baseline. We used Adam optimizer with learning rate 0.0005 and weight decay as $e^{-4}$. 

\begin{equation} \label{eq:lossCE}
L_{CE} = (-\sum^8_{c=1} \tilde{y}_{i=1}log(p^c(x_i,\theta))) ) 
\end{equation} 
Here $\theta$ refers to the model parameters $p^c$ refers to prediction probability of class c and $\tilde{y}_{c}$ refers to ground truth value of that class on a sample.

\subsection{Semi-Supervised learning with complementary labels}
There are a total of 1089929 images in train set, but nearly $50\%$ of this data, precisely 502970 images have invalid label(-1), therefore do not belong to any of the eight classes considered as part of the challenge. Therefore in total we have 586959 images as actual training images. To improve upon the performance of the fully supervised approach we could use these invalid images as unlabeled images and make use of semi-supervised learning which benefits from unlabeled images. Motivated from a recent work in semi-supervised learning named MutexMatch \cite{duan2022mutexmatch} which effectively uses unlabeled data to improve overall performance of the model in limited label setting.

MutexMatch uses a fixed threshold to divide unlabeled samples as high confident and low confident ones. Unlike other works in semi-supervised learning which emphasise on utilizing the high confident samples in varied varieties, MutexMatch uses the low confident samples to predict negative labels which is a simpler goal by the means of a True-Negative classifier(TNC). 

Similar to general approaches to semi-supervised learning it uses True-Positive classifier(TPC) to classify the images into considered set of classes. It has a supervised loss $L_{sup}$ \ref{eq:lossCE} on model predictions and ground truth labels, which helps the network learn features and weights, to classify appropriately. Once the model has learnt from the labeled data, this model is used to predict pseudo-labels on the unlabeled data using the True-Positive classifier. Fixed threshold is used to separate unlabeled samples into confident and non-confident samples. Predictions on confident samples are used as pseudo-labels to help the model utilize unlabeled data. 

To learn pseudo-labels effectively a pseudo-label loss $L_p$ is introduced which is the cross entropy$^{\ref{eq:lossCE}}$ loss between predictions of the model on weak and strong augmentations of high confident samples.  Two more losses $L_{sep}$ and $L_n$ involve the predictions from the complementary labels on low confident samples on the True-Negative classifier. $L_{sep}$ loss is cross entropy$^{\ref{eq:lossCE}}$ on TNC prediction of weak augmentation and the class with lowest confidence on weak augmented image as predicted by the TPC. $L_n$ $^{\ref{eq:lossLn}}$ is defined only on the low confident samples, which is negative consistency loss between prediction of TNC on weak and strong augmented versions of an image for top-k complementary predicted classes. 
 
\subsubsection{Loss functions}
\begin{equation} \label{eq:lossLp}
L_{p} = \frac{1}{\mu_B}\sum^{\mu_B}_{n=1}1(max(p^w_n) \geq \tau)H(\tilde{p}^{w}_{n},\tilde{p}^{s}_{n}))
\end{equation}
Here $\mu_B$ is the number of unlabeled images in a batch, H is the cross entropy function as in \ref{eq:lossCE}, $p^w_n$ is the TPC prediction on weak augmented image,  $\tilde{p}^{s}_{n}$ is the TPC prediction on strong augmented image, $\tilde{p}^{w}_{n}$ is $argmax(p^w_n)$. 

\begin{equation} \label{eq:lossLn}
L_{n} = \frac{1}{\mu_B}\sum^{\mu_B}_{n=1}1(max(p^w_n)<\tau)(-\frac{1}{k}\sum^C_{i=1} g_{n,(i)}r^{w}_{n,(i)}log(r^{s}_{n,(i)}))
\end{equation}
Here $\mu_B$ and $p^w_n$ are as defined above, $r^{w}_{n,(i)}$ is the ${i}^{th}$ probability component in the prediction of TNC on weak augmented image, $r^{s}_{n,(i)}$ is the $i^{th}$ probability component in the prediction of TNC on strong augmented image, $g_{n,(i)}$ is a mask which selects the classes that a in top-k of largest probability components in the complementary label prediction.

Inspired from \cite{AdaCM, xu2021dash, gera2023SSMFAR} which use dynamic adaptive threshold \ref{eq:thresh} that caters to inter and intra class differences that exist in facial expressions, we use dynamic adaptive threshold to divide the unlabeled samples into confident and non-confident ones. This class adaptive threshold is calculated by the model's performance on train set and is dynamically scaled up as the training progresses.

\begin{equation} \label{eq:thresh}
    T^c = \frac{\beta*({\frac{1}{N^s}}\sum^{N^s}_{i=1}\delta^c_i*p_i)}{1+\gamma^{-ep}} \hspace*{1em},\hspace*{1em} \mathrm{where}
\end{equation}
\begin{equation*} 
    \delta^c_i = \left \{ 
                        \begin{array}{ll}
                            1 & \mathrm{if \hspace*{1em}}  \tilde{y}_i = c, \\ \\
                            0 & \mathrm{otherwise}.
                        \end{array}
    \right.
\end{equation*}
The values  $\beta=0.95$ and $\gamma=e$ are taken from \cite{AdaCM}

\subsection{Noise aware model} 

During the training of the fully supervised model, we observed that accuracy and f1-score on an average was 98.7 and 98.2\% respectively on the training set, whereas average accuracy was 52.6 and the best f1-score was obtained to be 32.35\%.This shows that the model doesn't generalize well on unseen data. Generally, models do not generalize well due to the following reasons:
\begin{itemize}
    \item Capacity of the model : If the capacity of the model is low, then the model underfits the data leading to poor generalization. This can be solved by using a model with more capacity. But it can't be the case since the accuracy and f1-score on train is very high implying over fitting of model. 
    \item Distribution change : If there is a mismatch in the distribution of training data and the unseen validation data, then the model performs poorly on unseen data. But in general we treat the training data and unseen data are sampled from same distribution.
    \item Presence of noise : If there is noise in the labels of dataset  and if the model over-fits on the training data, then the model can't generalize well.  
\end{itemize}

 Assuming that there is noise in the labels, To deal with the noisy label problem, We propose a noise aware model.

This model has its backbone as resnet-18 followed by a fully connected layer with softmax activation to predict the 8 classes in expression classification task. We use pretrained weights from resnet18 model trained on affectnet dataset. In this model, we use two different types of augmentations of the image for consistency. MSE loss$^{\ref{eq:lossMSE}}$ is used for consistency and Weighted cross entropy loss$^{\ref{eq:weightedCE}}$ is used  for supervised loss. The weights are learnt by the model as per in SCN \cite{wang2020SCN} paper. Not all the samples are sent to supervised loss. We take out the samples whose prediction probabilities that fail to be greater than dynamic adaptive threshold{\cite{gera2022dynamic}}.

We use dynamic adaptive threshold in every epoch to tackle inter and intra class differences in prediction probabilities of expression classification. Dynamic adaptive threshold was calculated by taking the class vise mean of all the prediction probabilities in every batch. For a given class $i$, we treat the samples whose ground truth prediction probabilities that are greater than the dynamic adaptive threshold  for class $i$ as clean and those samples that fails are treated as noisy samples. Only clean samples are used for supervised loss and all the samples are used for consistency loss where we force the consistency between the attention maps of the weak augmented image and strong augmented image. 

The attention maps{\cite{lfa}} are calculated in the following way:
     
     We first extract the feature maps from last but second layer from the backbone and weights from the fully connected layer by multiplying the weights and feature maps we obtained the attention maps for every class. 
\begin{equation} \label{eq:weightedCE}
L_{WCE} = \frac{1}{N}(-\sum^N_{i=1} log(\frac{e^{\alpha_{i}W^{T}_{y_{i}}}x_{i}}{\sum^{C}_{j=1}e^{\alpha_{i}W^{T}_{j}}x_{i}})) 
\end{equation} 

\begin{equation} \label{eq:lossMSE}
L_{MSE} = -\frac{1}{NLHW}\sum^N_{i=1}\sum^L_{j=1} ||AM_{ij}-AM^{'}_{ij}||_{2}
\end{equation} 

We used Adam optimizer with 0.0001 as learning rate on all the model parameters.

\subsubsection{Loss functions}
We have used weighted cross entropy loss$^{\ref{eq:weightedCE}}$ for supervised loss where N is number of samples in batch C is number of classes and $\alpha_{i}$ are the weights.
The loss function that we have used for consistency loss is MSE loss$^{\ref{eq:lossMSE}}$ where N is number of samples in the batch, L is number of feature maps, H and W are height and width of feature maps, AM is attention maps on weak augmented image and $AM^{'}$ is attention maps on strong augmented flipped image.

\subsection{Problem formulation}
Let $D = {\{(x_i, \tilde{y}_i)\}}^N_{i=1}$ be the  dataset of N samples. Here $x_{i}$ is $i^{th}$ image where $\tilde{y}_i$ represents expression class ${y}^{Exp}_i)$ of $i^{th}$ image. The backbone network is parameterized by $\theta$ ( ResNet-18 \cite{he2016deepresnet} pre-trained on different datasets like rafdb \cite{li2017rafdb} and affectnet \cite{mollahosseini2017affectnet} for different approaches as backbone). We denote $x_w$ as weak augmented image and $x_s$ as strong augmented image, $P_{Exp}$ represent probability distribution predicted by Expression classifier. Weak augmentations include random cropping with padding and horizontal flipping of the input image. Strong augmentations includes weak augmentations along with Randaugment \cite{Randaug}.

\section{Dataset }

\subsection{Dataset}
s-AffWild2 \cite{zafeiriou2017affwild_dataset} database is a static version of Aff-Wild2 database and contains a total of  11,10,367 images for training and 4,53,535 images for validation. Out of which 5,02,970 images are to be disregarded from training set and 20,438 images are not present but the image paths are given. In validation data 2,79,749 are valid and remaining all are with -1 as label which we should disregard.

\subsection{Implementation Details}

The performance measure fo experssion classification problem is the average F1 score on all the 8 classes.The F1 score is a harmonic mean of the recall (i.e. Number of positive class images correctly identified out of true positive class) and precision (i.e. Number of positive class images correctly identified out of positive predicted). The F1 score takes values in the range [0, 1]. Here we need to present F1 score as percentage. The F1
score is defined as:
\begin{equation}
F_{1} = {\frac{2*precision*recall}{precision+recall}}
\end{equation}


\section{Results }
We report our results on the official validation set  from the ABAW 2023 Challenge \cite{kollias2023abaw} in Table \ref{tab:Tab1} . Our best performance achieves overall score of 33.46\% on validation set which is a significant improvement over baseline.

\begin{table}[hbt!]
\centering
    \caption{Performance comparison on Aff-Wild2 validation set}
    \begin{tabular}{c|c}
         \hline
         Method & Exp-F1 score  \\
         \hline
         \hline
         Baseline \cite{kollias2023abaw}  & \textit{23} \\
         Fully supervised model     &  \textit{32.34}   \\
         Semi-Supervised learning with complementary labels & \textit{32.82}\\
         Noise aware model & \textit{33.46}\\
         
         \hline
    \end{tabular}
    \label{tab:Tab1}
\end{table}

\section{Conclusions}
In this paper, we presented multiple methods. Firstly fully supervised model gave F1 score that is greater than the given baseline model{\cite{kollias2023abaw}} by 9.34\%. Then to overcome the limitations of fully supervised model, we proposed semi supervised learning with complementary labels and noise aware model that perform better over baseline by a margin of 9.38\% and 10.46\% respectively.


\section*{Acknowledgments}
We dedicate this work to Our Guru and Guide Bhagawan Sri Sathya Sai Baba, Divine Founder Chancellor of Sri Sathya Sai Institute of Higher Learning, Prasanthi Nilayam, Andhra Pradesh, India. 

\bibliographystyle{unsrt}
\bibliography{references}  

\end{document}